\documentclass[conference]{IEEEtran}
\IEEEoverridecommandlockouts

\usepackage{cite}
\usepackage{amsmath,amssymb,amsfonts}
\usepackage{algorithmic}
\usepackage{graphicx}
\usepackage{textcomp}
\usepackage{xcolor}
\def\BibTeX{{\rm B\kern-.05em{\sc i\kern-.025em b}\kern-.08em
    T\kern-.1667em\lower.7ex\hbox{E}\kern-.125emX}}
    
\usepackage{times}
\usepackage{soul}
\usepackage{url}
\usepackage[hidelinks]{hyperref}
\usepackage[utf8]{inputenc}
\usepackage[small]{caption}
\usepackage{graphicx}
\usepackage{amsmath}
\usepackage{amsthm}
\usepackage{booktabs}
\usepackage{algorithm}
\usepackage{algorithmic}
\usepackage{titling}
\usepackage{multicol}
\usepackage{amssymb}
\usepackage{dsfont}
\usepackage{mathtools}
\usepackage{graphicx}
\usepackage{xcolor}

\definecolor{mypink1}{rgb}{0.95, 0.35, 0.6}

\begin{document}

\title{Adversarial Training of Variational Auto-encoders for Continual Zero-shot Learning(A-CZSL)}

\author{{
Subhankar Ghosh}\\
\textit{Aerospace Engineering Department}\\
\textit{Indian Institute of Science}\\
Bengaluru, India\\
{subhankarg@alum.iisc.ac.in}
}

\date{}
\maketitle
\begin{abstract}
Most of the existing artificial neural networks(ANNs) fail to learn continually due to catastrophic forgetting, while humans can do the same by maintaining previous tasks' performances. Although storing all the previous data can alleviate the problem, it takes a large memory, infeasible in real-world utilization. We propose a continual zero-shot learning model(A-CZSL) that is more suitable in real-case scenarios to address the issue that can learn sequentially and distinguish classes the model has not seen during training. Further, to enhance the reliability, we develop A-CZSL for a single head continual learning setting where task identity is revealed during the training
process but not during the testing. We present a hybrid network that consists of a shared VAE module to hold information of all tasks and task-specific private VAE modules for each task. The model's size grows with each task to prevent catastrophic forgetting of task-specific skills, and it includes a replay approach to preserve shared skills. We demonstrate our hybrid model outperforms the baselines and is effective on several datasets, i.e., CUB, AWA1, AWA2, and aPY. We show our method is superior in class sequentially learning with ZSL(Zero-Shot Learning) and GZSL(Generalized Zero-Shot Learning). Our code is available at
\textcolor{mypink1}{{\url{https://github.com/CZSLwithCVAE/CZSL_CVAE}}}
\end{abstract}

\section{Introduction}
Conventional supervised machine learning (ML) and,
more recently, deep learning algorithms have shown remarkable performance on image classification, Computer
Vision and Natural Language Processing. Despite the recent success of supervised machine learning algorithms,
they have two significant limitations: 1) Conventional machine learning models have restricted themselves to training limited classes. When any example from the novel/unseen class occurs at the test time, such examples can not be correctly classified. 2) Conventional machine learning models cannot continually learn over time by accommodating new knowledge from streaming data while retaining learned information. We still need to improve our existing algorithms to achieve human-level performance the way humans learn \textcolor{blue}{\cite{a1}} sequentially without forgetting previous tasks throughout their life. Can we build an ANN model that can learn sequentially and simultaneously works for zero-shot learning(distinguish the classes it has not seen during training)? The name of such methods is continual zero-shot learning.

ZSL(zero-shot learning), where a trained model sees data only from unseen classes during testing, and GZSL(generalized zero-shot learning), where data may come from both seen and unseen classes for prediction. Both mentioned tasks are challenging for a model to perform continually. Researchers have addressed both ZSL\textcolor{blue}{\cite{a2, a3}} and continual learning \textcolor{blue}{\cite{a4, a14, a19}} approaches separately. Therefore, a more preferable and feasible
approach needs to tackle continual learning and unseen
object problems simultaneously. This model aims to leverage
the advantages of both continual
learning and  zero-shot learning  in a single framework. Consequently, we propose a model that can predict the unseen class object and
adapt to a new task without completely forgetting the knowledge about
the previous task.
References \textcolor{blue}{\cite{a5, a6}} combine ZSL with continual learning before, though our approach is entirely different from that, but is more realistic and beats their baselines results on the same datasets.\\
A traditional zero-shot learning(ZSL) method identifies unseen classes
using seen class samples and class embeddings. Despite showing a promising performance of traditional
ZSL, the method is unable to learn from streaming data. Generative
approaches have received quite an attention over embedding
based approaches due to the synthesized ability of
unseen class features and a significant improvement in the
model’s performance.
In this paper, we propose a novel adversarial training of variational autoencoders(VAEs) for continual zero-shot learning. Here, the model composes a task-specific private module that is learned for each task, and a task-invariant shared module that is learned for all tasks. Task-specific in a sense, the 1st private module learns from the 1st task's real data only, whereas the $t^{th}(t = 2,3,...,T)$ private module gets trained by real data from $t^{th}$ task and replay synthesized data from all previous (t-1) tasks generated from the shared module. We tackle both ZSL and continual learning together by using CVAE(conditional variational autoencoders) \textcolor{blue}{\cite{a11}} that transfer knowledge from seen to unseen classes through class embeddings\textcolor{blue}{\cite{a12}} to counter ZSL problems.  As visual data is not available during training time, knowledge transfer from seen to unseen classes is formed through side information that makes semantic relationships between classes and class-embeddings. Our approach is motivated by the fact that processing and synthesizing images are time taking for continual learning when the number of classes is high. Therefore, instead of images, we train and test our model with the features of the same images generated using a pre-trained model. Another thing that motivates us is that the human brain structure is complex and contains billions of neurons\textcolor{blue}{\cite{a13}}, so we may need to eventually make complicated networks in the coming future containing a huge number of neurons
to learn sequentially.The main contributions of this work are summarized as follows:
\begin{itemize}
\item We develop an generative replay-based and structure-based continual zero-shot learning method using CVAE.
\item The proposed method is developed for a single head setting that is more convenient to solve real case scenarios.
\item We present results for four ZSL benchmark datasets for continual zero-shot learning.
\item We propose two types of modules: one, task-invariant holds information for all tasks, another is task-specific. If there are T tasks, our proposed architecture consists of one task-invariant VAE and T task-specific VAEs.
\end{itemize}
\section{Releted Work}
\subsection{Continual learning}
The main problem in continual learning is catastrophic forgetting. McCloskey and Cohen first introduced the term catastrophic forgetting or catastrophic interference in the 1980s \textcolor{blue}{\cite{a31}}. They claimed that catastrophic interference is a fundamental limitation of
neural networks and a downside of its high generalization ability. While the cause of catastrophic
forgetting has not been studied analytically, it is known that the neural networks parameterize the
internal features of inputs, and training the networks on new samples causes alteration in already
established representations.
There are three types of continual learning to tackle forgetting: regularization-based, memory-based, and structure-based methods. 
\subsection*{Regularization methods}
Here, the learning capacity is fixed \textcolor{blue}{\cite{a14, a15}}, and continual learning is performed by penalizing a network's parameters. Researchers use a new regularizer for a novel method. In \textcolor{blue}{\cite{a15}}, the essential parameters are computed online. They keep track of how the loss function changes due to a specific parameter change and accumulate this information during training. There should be a weight importance process for parameters selection to prioritize parameters usage—the way elastic weight consolidation(EWC)\textcolor{blue}{\cite{a14}} gives importance to parameters based on the Fisher information matrix. The usages of these methods are limited because they cannot perform well for a large number of tasks.
\subsection*{Memory-based methods}
Methods in this category try to prevent forgetting by either storing \textcolor{blue}{\cite{a17, a18}} or synthesizing data from previous classes. The first one needs memory for rehearsal, whereas the latter is a generative model like GAN\textcolor{blue}{\cite{a20}} or VAE \textcolor{blue}{\cite{a11}}, or both synthesize data of previous tasks to perform pseudo-rehearsal. The number of examplers stored decreases with the increase in classes if the memory budget is limited. Researchers have recently proposed using a tiny memory\textcolor{blue}{\cite{a17}} to store a few examples per class for old tasks.
\subsection*{Structure-based methods}
The third approach to mitigate forgetting is structure-based methods\textcolor{blue}{\cite{a21}}. The size of a network grows with each task to prevent catastrophic forgetting. Previous tasks' performance is maintained by freezing the learned module while accommodating new tasks by augmenting the network with new modules. The computational cost for this method is inevitable if the number of tasks is high.
\subsection{Zero-shot learning and Generalized zero-shot learning}
Recently, zero-shot learning(ZSL)\textcolor{blue}{\cite{a2, a3}} has attracted a lot of attention because the model can distinguish unseen classes during testing. ZSL models are able to do so by transferring knowledge from seen to unseen levels through a semantic relationship between classes and their attributes. We can transform a ZSL problem into a supervised machine learning problem using generative models like GAN or VAE, or both. Once a generative model gets trained, it can synthesize data for unseen classes, and the data is useful for training a classifier like a conventional supervised problem. Another modification of ZSL is GZSL, a more practical approach, where data come from both seen and unseen classes.
\begin{figure}[H]
    \includegraphics[width = 10cm, height = 8cm]{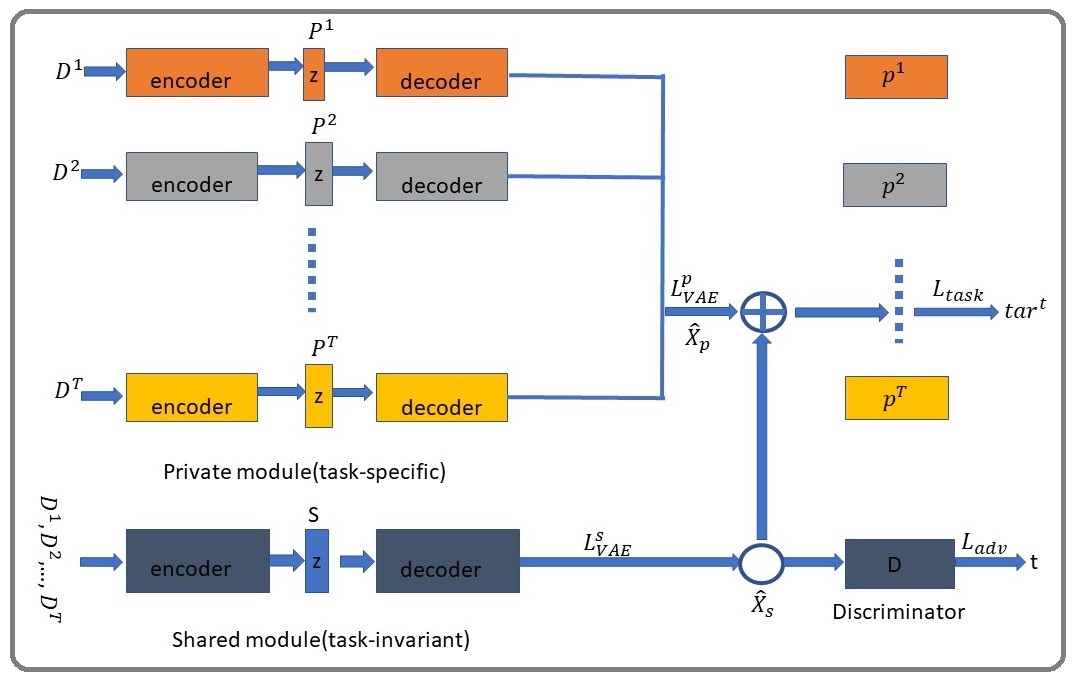}
    \caption{shows our model at training time where the shared module plays a minimax game with the discriminator to generate task-invariant $\hat{X}_s$ features. In contrast, the discriminator attempts to assign task labels to the synthesized features($\hat{X}_s$). The discriminator tries to maximize the probability of task label. Architecture grows because of the task-specific modules denoted as $P^t$ and $p^t$, task-specific perceptron networks assign class labels during each task's training. To prevent forgetting shared module is trained with replay examples of previously learned classes generated from the decoder of the shared module using $z^{'} \sim \mathcal{N}(\mu = 0, \sum = 1)$.}
\end{figure}

\begin{figure}[hbt!]
    \includegraphics[width = 9cm, height = 7cm]{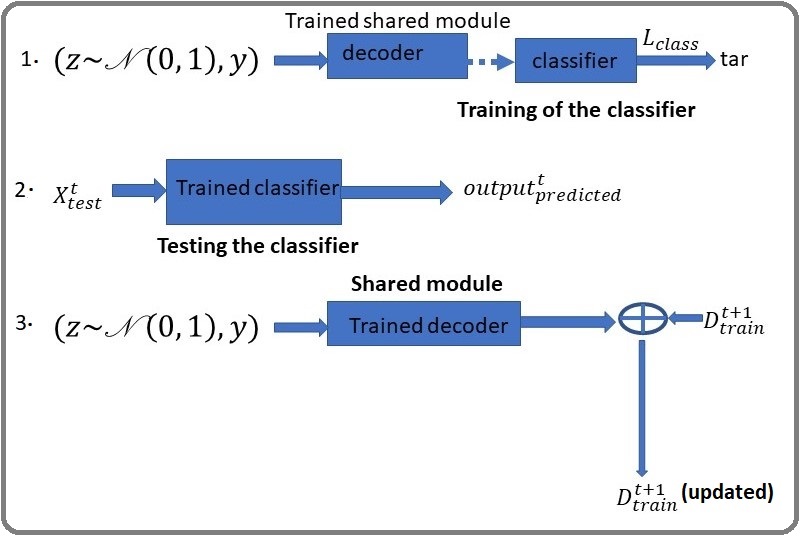}
    \caption{ has three parts:\\
1) \textbf{Training of the classifier}(separate) with generated data(of both seen and unseen classes) using the shared module's trained decoder.\\
2) \textbf{Testing the classifier} for both seen and unseen classes(ZSL).\\
3) The \textbf{shared module}'s decoder is used to generate replay samples of previous tasks that get concatenated to the next task to update the model with new and old tasks.
}
\end{figure}
\subsection{Continual Zero-shot Learning}
    Reference \textcolor{blue}{\cite{a6}} proposed a continual zero-shot learning model called Generalized continual zero-shot learning(GCZSL), a single head CZSL where the task identity is revealed during training but not during testing. To mitigate forgetting, they used knowledge distillation and stored few actual data per class using a small episodic memory. The GCZSL and reference \textcolor{blue}{\cite{a5}} used the single-head setting, where the task identity is disclosed during training but not at test time. There are a few works\textcolor{blue}{\cite{a29}, \cite{a30}} that used the multi-head setting. Reference \textcolor{blue}{\cite{a29}} proposed an average gradient episodic memory(A-GEM) based CZSL, and \textcolor{blue}{\cite{a30}} offered a generative model-based CZSL.
\subsection{Adversarial learning}
Adversarial learning has usages in many domains such as generative models\textcolor{blue}{\cite{a20}}, object composition\textcolor{blue}{\cite{a22}}, representation learning\textcolor{blue}{\cite{a23}}, domain adaptation\textcolor{blue}{\cite{a24}}, active learning\textcolor{blue}{\cite{a25}} etc. In adversarial training, a model learns the parameters through the minimax game, where a module wants to maximize the cost function, and another wants to minimize the same. This paper shows shared play the minimax game with discriminator, where shared tries to minimize the loss function, and the discriminator wants to maximize.
\section{Adversarial Training of Conditional Variational Autoencoders for Continual Zero-Shot Learning(A-CZSL)}
We study the problem of learning a sequence of T data distributions denoted as $D_{tr} = \{D^1_{tr}, D^2_{tr}, ..., D^T_{tr}\}$, where $D^t_{tr} = \{(X^t_i, Y^t_i, tar^t_i, T^t_i)^{n_t}_{i = 1}\}$ is the data distribution for the task t with $n_t$ sample tuples of input($X^t \in \mathcal{X}$), target label($tar^t \in tar$), attributes of classes ($Y^t \in \mathcal{Y}$), and task label($T^t \in \mathcal{T}$). $D^t_{tr}$ contains seen class information. Apart from this, class embeddings of unseen classes($\mathcal{U}_c = \{(a_i)^{n_{uc}}_{i = 1}\}$) are also available, where $n_{uc}$ is the number of unseen labels. The goal is to learn a sequential function, $f: (z\sim\mathcal{N}(0, 1), \mathcal{Y}) \rightarrow \hat{\mathcal{X}}_s$, for each task, where $\hat{\mathcal{X}}_s$ is synthesized data generated from the shared module. The synthetic data can be used to train a supervised classifier. We aim to learn another function, $f_{\theta_c}: \hat{\mathcal{X}} \rightarrow tar$, after training each task, that can map input(from seen or unseen or both classes) into it's target output without affecting the previous model's performance on prior works. We try to achieve our goal by training two separate modules: shared and private, to enhance a better knowledge transfer from seen to unseen classes and better forget avoidance of prior knowledge. The model prevents catastrophic forgetting in shared and private spaces separately and begins learning $f_{\theta}^t$ where $\theta \in (\theta_S, \theta_P)$ as mapping function from $(\mathcal{X}^t_{tr}, \mathcal{Y}^t_{tr})$ to $tar^{t}$. We use some n samples per class to be synthesized prior to $t^{th}$ task and accumulate the generated data to the current task($t^{th}$) to train the model.
\begin{center}
    $D^{t}_{tr} \leftarrow D^{t}_{tr}\cup D_{gen}^{1:(t-1)}$
\end{center}
The cross-entropy loss function for the $f^t_{\theta}$ mapping corresponds to:
\begin{multline}
    L_{task = t}(f^t_{\theta}, D^t_{tr}) = -\mathop{\mathbb{E}}_{(\mathcal{X}^t, \mathcal{Y}^t, tar^t) \sim D^t}\\
    \left[\sum_{c = 1}^C\mathds{1}_{(c = tar^t)}log(\sigma(f_{\theta}^t(\mathcal{X}^t, \mathcal{Y}^t)))\right]
\end{multline}
Where $\sigma$ is the softmax function, in learning a sequence of tasks, an ideal $f^t$ maps the input features $X^t$ to two independent feature spaces: $X^t_s$ a shared features space among all tasks and $X^t_p$ remains private for each task. Both $X^t_s$ and $X^t_p$ get concatenated and fed to a task-specific multi-layer perceptron network to get desired output labels.

We introduce a mapping named shared ($S_{\theta_s}:X\rightarrow \hat{X}_s$) and train it to generate features by feeding noise into the shared module's decoder to fool a discriminator D. In contrast, the D($D_{\theta_d}:\hat{X}_s\rightarrow T$) try to assign the synthesized features to their corresponding task labels($T^{t\in \{0,1,2,..,T\}}$). The decoder and the discriminator can do so when the D gets trained to maximize the probability of assigning correct task labels to the features generated from the shared module. Simultaneously, the shared tries to minimize the same probability.

The corresponding cross-entropy adversarial loss for the minimax game:
\begin{multline}
    L_{adv}(D, S, D^t_{tr}) = \mathop{min}_S \mathop{max}_D\\
    \sum^T_{t = 0}\mathds{1}_{t = t^t}\left[log(D(S(\mathcal{X}^t, \mathcal{Y}^t)))\right]
\end{multline}
The extra-label zero is there for fake data generated from the Gaussian distribution with mean = 0 and std = 1. In most cases, we use adversarial training in a generative adversarial network that tries to learn the input data distribution in order to synthesize more data from the same distribution. Here we do the same by utilizing generative models task-invariant shared(VAE), and task-specific private(VAE); both try to learn input data distribution.

To facilitate adversarial training for S, we use the Gradient Reversal Layer \textcolor{blue}{\cite{a28}} that directly tries to maximize the discriminator's loss. The layer acts like an identity function during forward-propagation but multiplies the loss with a negative one during backpropagation in order to maximize the cost function for the discriminator. The adversarial training between the discriminator and the shared is complete when the discriminator can no longer predict the correct task label for features generated from the shared module. The private module, however, merely learns any task-invariant features.

\subsection*{Variational autoencoders}
Autoencoders can effectively learn feature space and representation\textcolor{blue}{\cite{a15, a22}}. A variational Autoencoder(VAE) is a generative model that follows an encoder-latent vector-decode architecture of classical autoencoder, which places a prior distribution on the feature space and uses an expected lower bound to optimize the learned posterior. Conditional VAE is an extension of the VAE, where data are fed to network with class properties such as labels, attributes, etc. The VAE is a fundamental building block of our approach. Variational distribution aims to find a true conditional probability distribution over the latent variables z through minimizing their distance using a variational lower bound limit. The loss function for a VAE is:
\begin{multline}
    L_{VAE} = \mathop{\mathbb{E}}_{q_{\phi}({z}|{x})}\left[log(p_\theta({x}|{z}))\right] - D_{KL}(q_\phi({z}|{x})\parallel p_{\theta(z)})
\end{multline}
Where the first term is the reconstruction loss, and the second one is the KL divergence between $q({z}|{x})$ and p(z). The encoder predicts $\mu$ and $\sum$ such that $q_{\phi}({z}|{x}) = \mathcal{N}(\mu, \sum)$, from which a latent vector is synthesized via reparametrization process.

The final objective function of the model for the $t^{th}$ task is: 
\begin{multline}
L^{(t)} = \lambda_1 L_{adv} + \lambda_2 L_{task} + \lambda_3 L^s_{VAE} + \lambda_4 L^p_{VAE}
\end{multline}
Where, $\lambda_1,\lambda_2, \lambda_3,$ and $\lambda_4$ are regularizers to control the effect of each component. The full algorithm of the model is given in Algorithm \textcolor{blue}{[\ref{alg:algorithm}]}. 
\subsection{Avoid forgetting}
Catastrophic forgetting occurs because of the imbalance between old and new classes that results in a bias of the network towards the newest ones. One insight of our approach is to decouple the single representation learned for all tasks continually into two parts: private and shared. Though knowledge is transferred for ZSL and GZSL mostly from the shared module from seen to unseen classes. The critical approach is experience replay that gets concatenated to the current task's data during training of the model with the same task to avoid forgetting sequentially.
\subsection{Datasets}
We evaluate our model on four benchmark datasets used for ZSL: Attribute Pascal and Yahoo(aPY)\textcolor{blue}{\cite{a2}}, Animals With Attributes(AWA1, AWA2)\textcolor{blue}{\cite{a2}}, and Caltech-UCSD-Birds 200-2011(CUB)\textcolor{blue}{\cite{a26}}. Statistics of the datasets are presented in Table \textcolor{blue}{[\ref{table:1}]}.

\subsection{Continual Zero-shot learning(CZSL) setting}
The dataset we use follows the setting used in\textcolor{blue}{\cite{a5}}. It explains whether a class is seen or unseen is decided based on the number of tasks a model has been trained so far. If a model goes trained continually up to the $t^{th}$ task, the classes are assumed to be seen till the $t^{th}$ task, and the rest of the whole dataset's classes are accepted unseen for the model while training. 
\subsection{evaluation matrices}
We evaluate the resulting model on all previous tasks similar to\textcolor{blue}{\cite{a16, a18}} after training for each new task. We use ACC as the average test classification accuracy across all classes for GZSL, seen classes, and unseen classes for GSL to measure our model's performance. To measure forgetting, we calculate backward transfer, BWT that says how much learning new tasks have influenced previous tasks' performance. While $BWT>0$ indicates catastrophic forgetting and $BWT<0$, learning new tasks has helped improve performance on previous tasks. We calculate forgetting measure for seen classes only.
\begin{equation}
    BWT = \frac{1}{T-1}\sum^{T-1}_{t = 1}\left[R_{t,t}^{seen} - R_{T, t}^{seen}\right]
\end{equation}
\begin{equation}
    mSA = \frac{1}{T}\sum^{T}_{t = 1}R_{t, t}^{seen}
\end{equation}
mSA is the mean seen classification accuracy across all tasks.
\begin{equation}
    mUA = \frac{1}{T-1}\sum^{T-1}_{t = 1}R^{unseen}_{t, t}
\end{equation}
mUA is the measure of zero-shot learning performance for the model.
\begin{equation}
    mOA = \frac{1}{T}\sum^{T}_{t = 1}R_{t, t}^{overall}
\end{equation}
mOA is the measure of generalized zero-shot learning performance.
\begin{equation}
    mH = \frac{1}{T-1}\sum^{T-1}_{t = 1}\left[\frac{2*{R^{seen}_{t,t}}*R^{unseen}_{t,t}}{R^{seen}_{t,t}+R^{unseen}_{t,t}}\right]
\end{equation}
mH is the hermonic mean classification accuracy. 

\begin{algorithm}[H]
\caption{Continual Zero-shot Learning}
\label{alg:algorithm}
\textbf{Input}: $(\mathcal{X}, \mathcal{Y}, tar) \sim D^{all}$\\
\textbf{Parameters}: $\theta_S, \theta_P, \theta_D, \theta_c$\\
\textbf{Output}: $\hat{X}_S, \hat{X}_P$
\begin{algorithmic}[1] 

\STATE $D^{gen} \leftarrow \{\}$
\FOR{t $\leftarrow$ 1 to T}
  \FOR{e $\leftarrow$ 1 to epochs}
    \FOR{k $\leftarrow$ 1 to $S_{steps}$}
      \STATE Compute $L_{task}$ using $(\mathcal{X}^t, \mathcal{Y}^t, tar^t) \in D^t$ 
      \STATE Compute $L_{adv}$ using  $(\mathcal{X}^t, \mathcal{Y}^t, t) \in D^t$
      \STATE Compute $L_{VAE}^S$ for shared module using $(\mathcal{X}^t, \mathcal{Y}^t)\in D^t$
      \STATE Compute $L_{VAE}^P$ for private module using $(\mathcal{X}^t, \mathcal{Y}^t)\in D^t$
      \STATE $L^{(t)} = \lambda_1 L_{adv} + \lambda_2 L_{task} + \lambda_3 L^s_{VAE} + \lambda_4 L^p_{VAE}$
      \STATE $\theta_S^{'} \leftarrow \theta_S - \alpha_S \nabla L^{(t)}$
      \STATE $\theta_P^{'} \leftarrow \theta_P - \alpha_P \nabla L^{(t)}$
    \ENDFOR
    \FOR{j $\leftarrow$ 1 to $D_{steps}$}
      \STATE Compute $L_{adv}$ for D using ($S(x)^t, tar^t$) and ($z^{'} \sim \mathcal{N}(\mu = 0, \sum = 1), tar = 0$)
      \STATE $\theta_D^{'} \leftarrow \theta_D - \alpha_D \nabla L^{(t)} $ 
    \ENDFOR
  \ENDFOR
  \STATE Generate data from the shared module for seen and unseen classes to train a separate classifier.
  \STATE $D \leftarrow D_{seen} \cup D_{unseen}$
  \FOR{$C_e \leftarrow$ 1 to $C_{epochs}$}
  \STATE Compute $L_{class}$ using ($\mathcal{X}, tar$) $\in$ D
  \STATE $\theta_{c}^{'} \leftarrow \theta_c - \alpha_c \nabla L_{class}$
  \ENDFOR
  \STATE Test the classifier for seen data. 
  \STATE Test the classifier for unseen data(ZSL).
  \STATE Test the classifier for all seen and unseen data(GZSL).
  \FOR{c $\leftarrow$ 1 to C} 
  \STATE C is the replay classes.
    \FOR{i $\leftarrow$ 1 to n} 
    \STATE n is the number of samples to be generated per class for the experience replay.
    \STATE $(\mathcal{X}_i, \mathcal{Y}_i, tar_i) \sim D^{gen}$
      
    \ENDFOR
  \ENDFOR
  \STATE $D^{t+1} \leftarrow D^{t+1}\cup D^{gen}$

\ENDFOR
\end{algorithmic}
\end{algorithm}

Where $R_{j,i}$ is the test classification accuracy on task i after sequentially finishing learning the $j^{th}$ task.
\begin{table}
 \begin{tabular}{||c c c c c||} 
 \hline
 Dataset & Semantic Dim & \#Images & \#SC & \#UC \\ [0.5ex] 
 \hline\hline
 CUB & 312 & 11788 & 150 & 50 \\ 
 \hline
 aPY & 64 & 15339 & 20 & 12\\
 \hline
 AWA1 & 85 & 30475 & 40 & 10 \\
 \hline
 AWA2 & 85 & 37322 & 40 & 10\\ [1ex] 
 \hline
\end{tabular}
\caption{Datasets and their statistics, Where SC and UC are seen and unseen classes respectively.}
\label{table:1}
\end{table}
\section{Experiments}
This section consists of baselines and implementation
details we used in our experiment.

\subsection{Baselines}
The research on continual zero-shot learning(CZSL) has been less explored. References\textcolor{blue}{\cite{a5, a6}} has investigated the work before on a single-head setting that we represent in this paper. Reference \textcolor{blue}{\cite{a6}} used the following baselines, so we do the same in this paper.
\begin{itemize}
\item \textbf{AGEM} + \textbf{CZSL}\textcolor{blue}{\cite{a5}}: It is an average gradient episodic memory-based continual zero-shot learning. The authors of \textcolor{blue}{\cite{a5}} have mentioned the harmonic mean of the CUB dataset only. 
\item \textbf{SEQ} + \textbf{CVAE}\textcolor{blue}{\cite{a6}}: The authors train CVAE sequentially without considering any continual learning strategy. After SEQ+CVAE is trained on the current task, synthetic are generated using noise and class embeddings for all classes to train a separate classifier.
\end{itemize}
\subsection{Other Methods}
We also compare our results with CZSL-CV+mof\textcolor{blue}{\cite{a6}}, CZSL-CV+rb\textcolor{blue}{\cite{a6}}, and CZSL-CV+res\textcolor{blue}{\cite{a6}}.

\subsection{Implementation Details}
We use Pytorch as our framework. We train our model with a hundred epochs and a classifier for thirty epochs for each task on all datasets except CUB. We use the same number of epochs for the CUB dataset till task fifteen and then reduce to fifty for the model and ten for the classifier till task eighteen and again decrease to twenty for the model and five for the classifier. The Adam\textcolor{blue}{\cite{a27}} optimizer is used in all experiments, and the learning rate for the classifier and others is 0.0001 and 0.001, respectively. We use weight decay 0.0001 as a regularizer for the classifier. We use 500 hidden units for both shared and private modules. Latent dimension is 50, and batch size for both model and classifier is 61. We take $\lambda_1 = \lambda_2 = \lambda_3 = 1$, and $\lambda_4 = 0.5.$

\section{Results and Discussion}
In the first set of experiments, we measure mSA, mUA,
mH, and compare it against state-of-the-art
methods on CUB, aPY, AWA1, and AWA2 datasets.
\subsection{Performance on CUB Dataset}
\begin{figure}[H]
    \includegraphics[width = 9cm, height = 5cm]{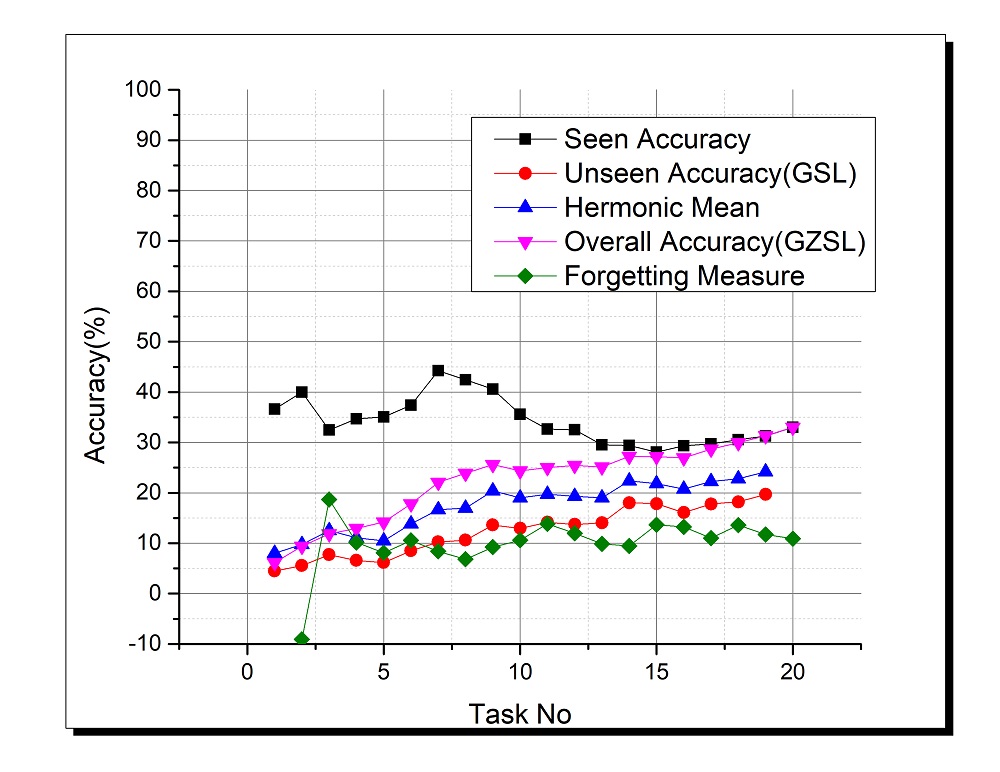}
    \caption{Results without adversarial training for the CUB dataset.}
    \label{fig:b3}
    \includegraphics[width = 9cm, height = 6cm]{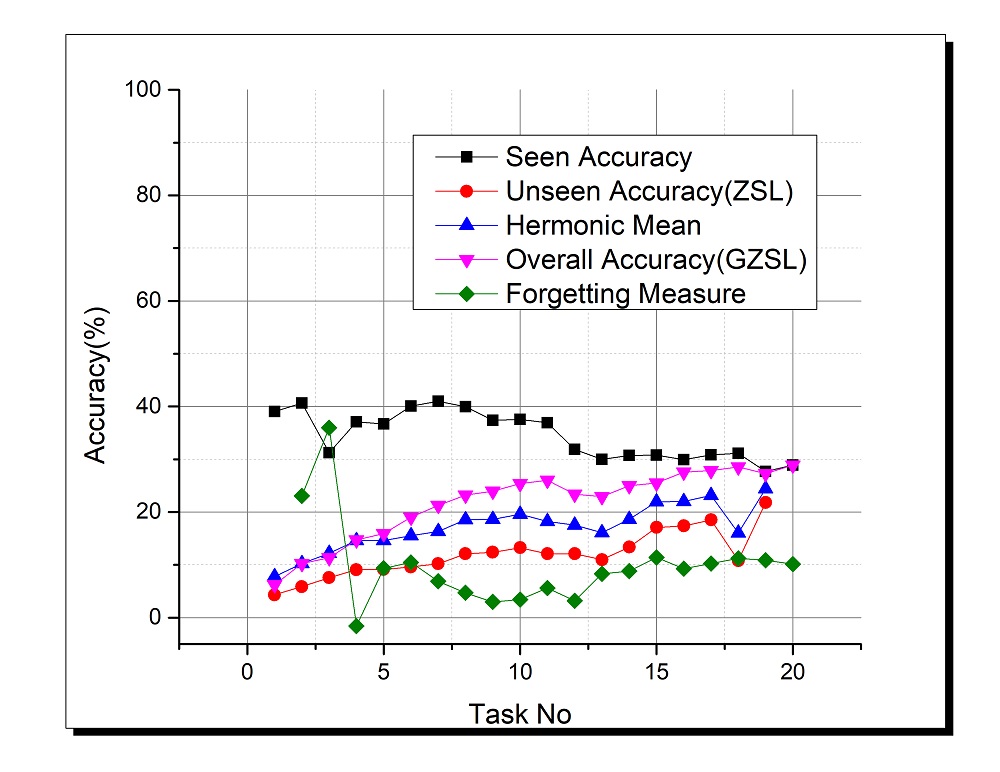}
    \caption{Results with adversarial training for the CUB dataset.}
    \label{fig:b4}
\end{figure}

\begin{table}
\begin{tabular}{ |p{2.5cm}|p{0.6cm}|p{1.4cm}|p{0.6cm}|p{1.7cm}|  }
 \hline
 \multicolumn{5}{|c|}{CUB} \\
 \hline
 Methods & mSA & mUA(ZSL) & mH & mOA(GZSL)\\
 \hline
 AGEM+CZSL\textcolor{blue}{\cite{a5}}   & -    &-&   13.20& -\\
 Seq-CVAE\textcolor{blue}{\cite{a6}}&   24.66  & 8.57   &12.18 & -\\
 \hline
 CZSL-CV+mof\textcolor{blue}{\cite{a6}}&43.73 & 10.26&  16.34& -\\
 CZSL-CV+rb\textcolor{blue}{\cite{a6}}&42.97 & 13.07&  19.53& -\\
 CZSL-CV+res\textcolor{blue}{\cite{a6}}&   \textbf{44.89}  & \textbf{13.45}&\textbf{20.15} & -\\
 \hline
 \multicolumn{5}{|c|}{ours} \\
 \hline
 A-CZSL$^{**}$  & 34.25   &12.42&17.41 & \textbf{22.40}\\
 A-CZSL$^{*}$& 34.47  & 12.00&17.15& 21.72\\
 \hline
\end{tabular}
\caption{Results for the CUB dataset, where mSA: Mean Seen Accyracy, mUA: Mean Unseen Accuracy, mH: Hermonic Mean Accuracy, mOA = Mean Overall Accuracy. The best results in the table are presented in boldface. ($^{**}$) and ($^{*}$) denote that the model has been trained without and with adversarially, respectively.}
\label{table:2}
\end{table}
We divide the CUB dataset into 20 tasks, where each consists of ten classes. We compare our results with several methods in Table \textcolor{blue}{[\ref{table:2}]}. Generalized CZSL\textcolor{blue}{\cite{a6}} used memory to store real data as the replay, and it achieved mH = 20.15. In contrast, our model achieves mH = 17.41 when each task gets trained for 100 epochs till the $15^{th}$ task, then from task $16^{th}$ to $18^{th}$ for 50 epochs and rest of the tasks for 20 epochs. Similarly, the classifier gets trained for 30 epochs till the task $15^{th}$, then from task $16^{th}$ to $18^{th}$ for 10 epochs and rest of the tasks for 5 epochs. The performance of our model per task is given in Figures \textcolor{blue}{[\ref{fig:b3}, \ref{fig:b4}]}.

\subsection{Performance on aPY Dataset}
\begin{figure}[H]
    \includegraphics[width = 9cm, height = 5cm]{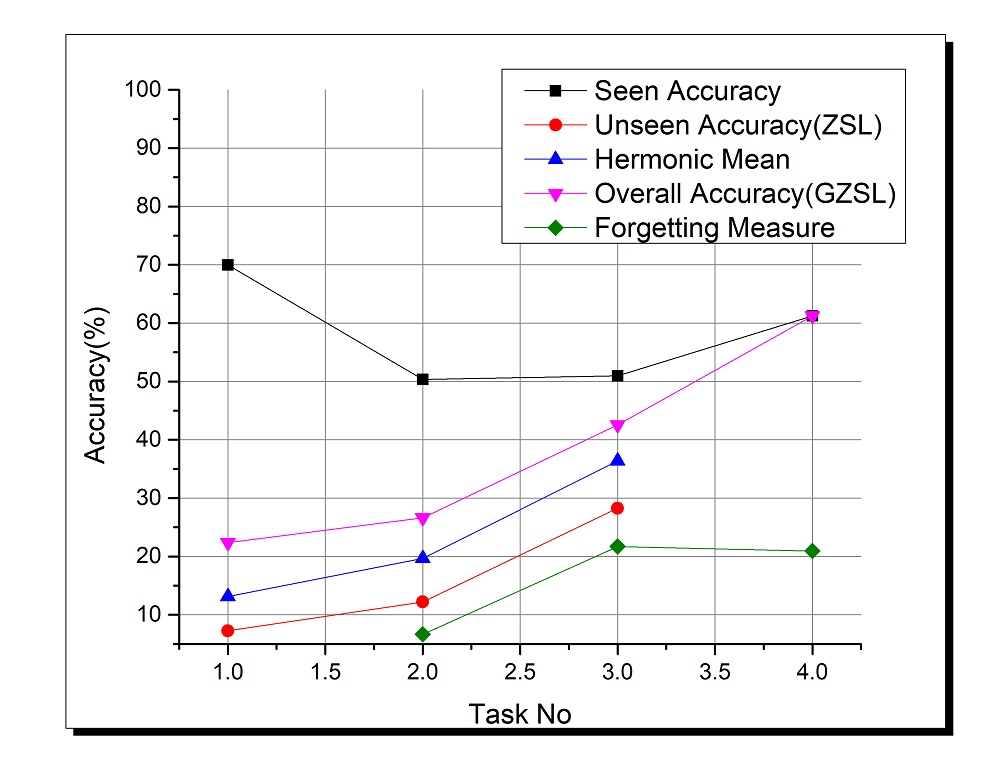}
    \caption{Results without adversarial training for the aPY dataset.}
    \label{fig:b5}
    \includegraphics[width = 9cm, height = 5cm]{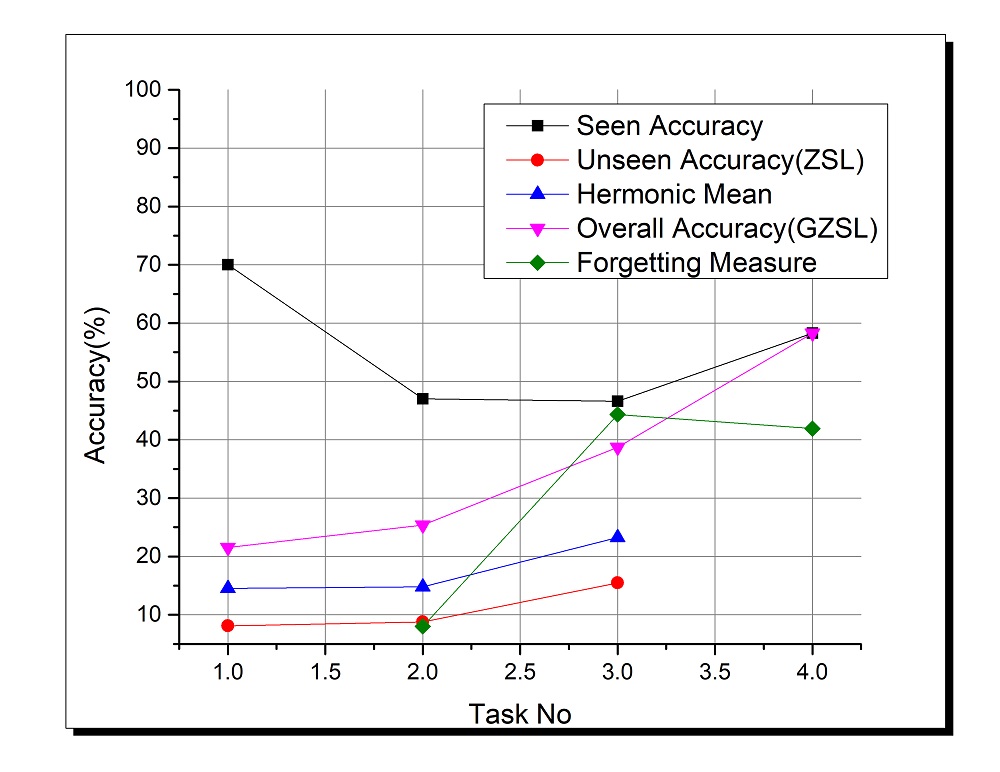}
    \caption{Results with adversarial training for the aPY dataset.}
    \label{fig:b6}
\end{figure}

\begin{table}
\begin{tabular}{ |p{2.5cm}|p{0.6cm}|p{1.4cm}|p{0.6cm}|p{1.7cm}|  }
 \hline
 \multicolumn{5}{|c|}{aPY} \\
 \hline
 Methods & mSA & mUA(ZSL) & mH & mOA(GZSL)\\
 \hline
 AGEM+CZSL\textcolor{blue}{\cite{a5}}   & -    &-&   -& -\\
 Seq-CVAE\textcolor{blue}{\cite{a6}}&   51.57&11.38&18.33& -\\
 \hline
 CZSL-CV+mof\textcolor{blue}{\cite{a6}}&64.91&10.79&18.27& -\\
 CZSL-CV+rb\textcolor{blue}{\cite{a6}}&64.45&11.00&18.60& -\\
 CZSL-CV+res\textcolor{blue}{\cite{a6}}&  \textbf{64.88}&15.24&\textbf{23.90}& -\\
 \hline
 \multicolumn{5}{|c|}{ours} \\
 \hline
 A-CZSL$^{**}$  &58.14&\textbf{15.91}&23.05&\textbf{38.20}\\
 A-CZSL$^{*}$& 55.46&11.2&18.63&35.97\\
 \hline
\end{tabular}
\caption{Results for the aPY dataset, where mSA: Mean Seen Accyracy, mUA: Mean Unseen Accuracy, mH: Hermonic Mean Accuracy, mOA = Mean Overall Accuracy. The best results in the table are presented in boldface. ($^{**}$) and ($^{*}$) denote that the model has been trained without and with adversarially, respectively.}
\label{table:3}
\end{table}
We divide the aPY dataset into 4 tasks, where each consists of eight classes. We compare our results with several methods in Table \textcolor{blue}{[\ref{table:3}]}. Generalized CZSL\textcolor{blue}{\cite{a6}} used memory to store real data as the replay, and it achieved mH = 23.90. In contrast, our model achieves mH = 23.05 when each task gets trained for 100 epochs. Similarly, the classifier gets trained for 30 epochs for all tasks. The performance of our model per task is given in Figures \textcolor{blue}{[\ref{fig:b5}, \ref{fig:b6}]}.

\subsection{Performance on AWA1 Dataset}
\begin{table}[hbt!]
\begin{tabular}{ |p{2.5cm}|p{0.6cm}|p{1.4cm}|p{0.6cm}|p{1.7cm}|  }
 \hline
 \multicolumn{5}{|c|}{AWA1} \\
 \hline
 Methods & mSA & mUA(ZSL) & mH & mOA(GZSL)\\
 \hline
 AGEM+CZSL\textcolor{blue}{\cite{a5}}   & -    &-&   -& -\\
 Seq-CVAE\textcolor{blue}{\cite{a6}}&   59.27&18.24&27.14& -\\
 \hline
 CZSL-CV+mof\textcolor{blue}{\cite{a6}}&76.77&19.26&30.46& -\\
 CZSL-CV+rb\textcolor{blue}{\cite{a6}}&77.85&21.90&33.64& -\\
 CZSL-CV+res\textcolor{blue}{\cite{a6}}&   \textbf{78.56}&23.65&35.51& -\\
 \hline
 \multicolumn{5}{|c|}{ours} \\
 \hline
 A-CZSL$^{**}$  & 67.98&20.58&30.83& 43.08\\
 A-CZSL$^{*}$& 71.00&\textbf{24.26}&\textbf{35.75}&\textbf{50.9}\\
 \hline
\end{tabular}
\caption{Results for the AWA1 dataset, where mSA: Mean Seen Accyracy, mUA: Mean Unseen Accuracy, mH: Hermonic Mean Accuracy, mOA = Mean Overall Accuracy. The best results in the table are presented in boldface. ($^{**}$) and ($^{*}$) denote that the model has been trained without and with adversarially, respectively.}
\label{table:4}
\end{table}

\begin{figure}
    \includegraphics[width = 9cm, height = 5cm]{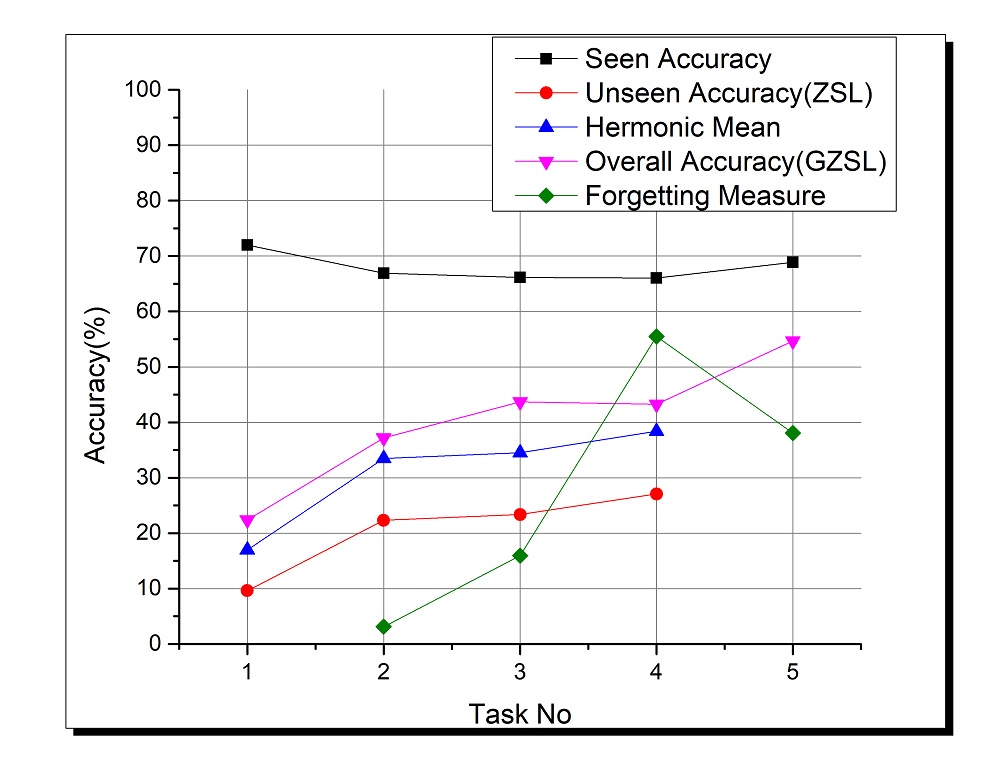}
    \caption{Results without adversarial training for the AWA1 dataset.}
    \label{fig:b7}
    \includegraphics[width = 9cm, height = 5cm]{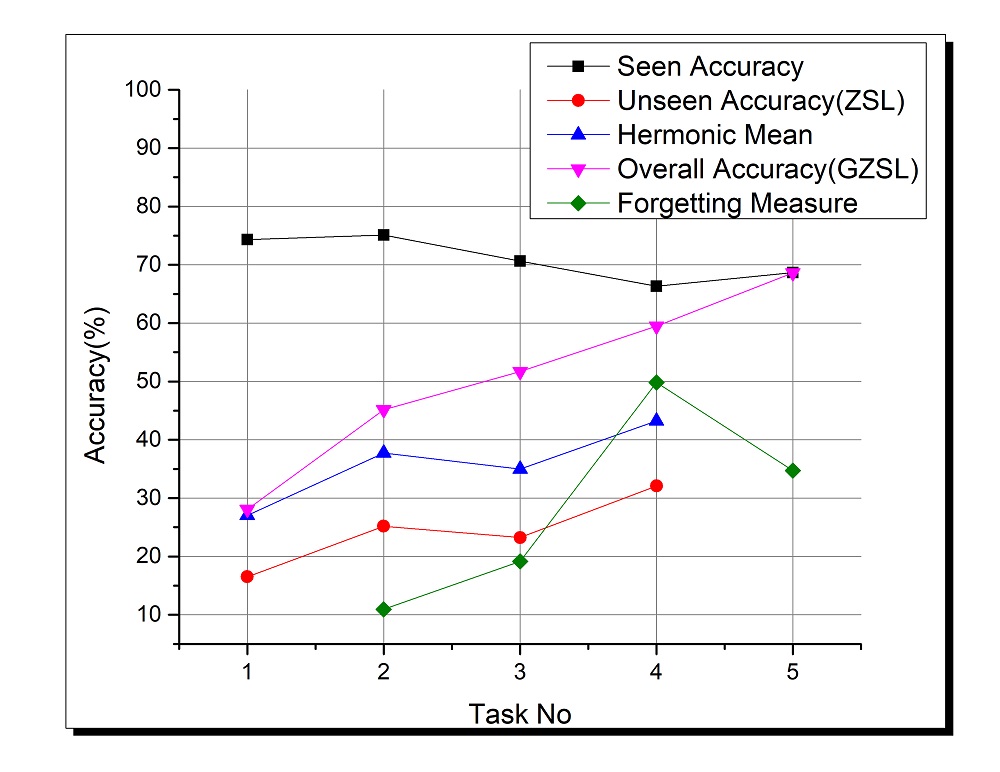}
    \caption{Results with adversarial training for the AWA1 dataset.}
    \label{fig:b8} 
\end{figure}

We divide the AWA1 dataset into 5 tasks, where each consists of ten classes. We compare our results with several methods in Table \textcolor{blue}{[\ref{table:4}]}. Generalized CZSL\textcolor{blue}{\cite{a6}} used memory replay to alleviate the catastrophic forgetting and used memory to store real data, and it achieved mH = 35.51. In contrast, our model achieves mH = 35.75 when each task gets trained for 100 epochs. Similarly, the classifier gets trained for 30 epochs for all tasks. The performance of our model per task is given in Figures \textcolor{blue}{[\ref{fig:b7}, \ref{fig:b8}]}.

\subsection{Performance on AWA2 Dataset}
\begin{table}[hbt!]
\begin{tabular}{ |p{2.5cm}|p{0.6cm}|p{1.4cm}|p{0.6cm}|p{1.7cm}|  }
 \hline
 \multicolumn{5}{|c|}{AWA2} \\
 \hline
 Methods & mSA & mUA(ZSL) & mH & mOA(GZSL)\\
 \hline
 AGEM+CZSL\textcolor{blue}{\cite{a5}}   & -    &-&   -& -\\
 Seq-CVAE\textcolor{blue}{\cite{a6}}&   61.42&19.34&28.67& -\\
 \hline
 CZSL-CV+mof\textcolor{blue}{\cite{a6}}&79.11&24.41&36.60& -\\
 CZSL-CV+rb\textcolor{blue}{\cite{a6}}&80.92&24.82&37.32& -\\
 CZSL-CV+res\textcolor{blue}{\cite{a6}}&   \textbf{80.97}&25.75&\textbf{38.34}& -\\
 \hline
 \multicolumn{5}{|c|}{ours} \\
 \hline
 A-CZSL$^{**}$  &70.05&22.85&32.98&44.97\\
 A-CZSL$^{*}$& 70.16&\textbf{25.93}&37.19&\textbf{51.55}\\
 \hline
\end{tabular}
\caption{Results for the AWA2 dataset, where mSA: Mean Seen Accyracy, mUA: Mean Unseen Accuracy, mH: Hermonic Mean Accuracy, mOA = Mean Overall Accuracy. The best results in the table are presented in boldface. ($^{**}$) and ($^{*}$) denote that the model has been trained without and with adversarially, respectively.}
\label{table:5}
\end{table}

\begin{figure}[H]
    \includegraphics[width = 10cm, height = 5cm]{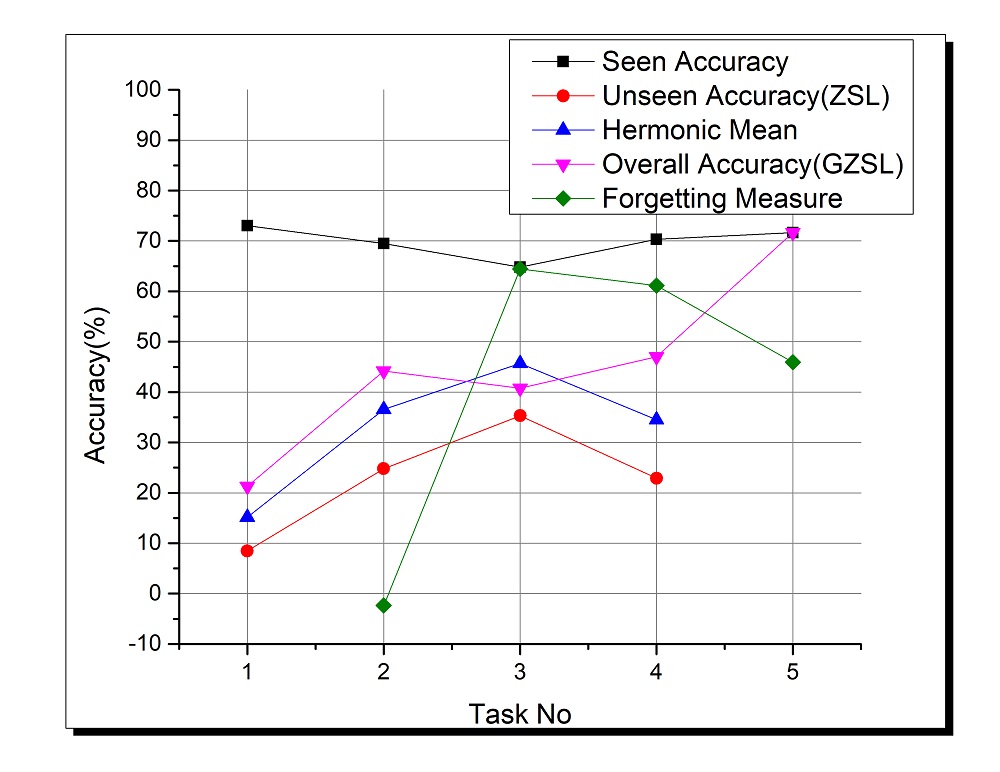}
    \caption{Results without adversarial training for the AWA2 dataset.}
    \label{fig:b9}
    \includegraphics[width = 10cm, height = 5cm]{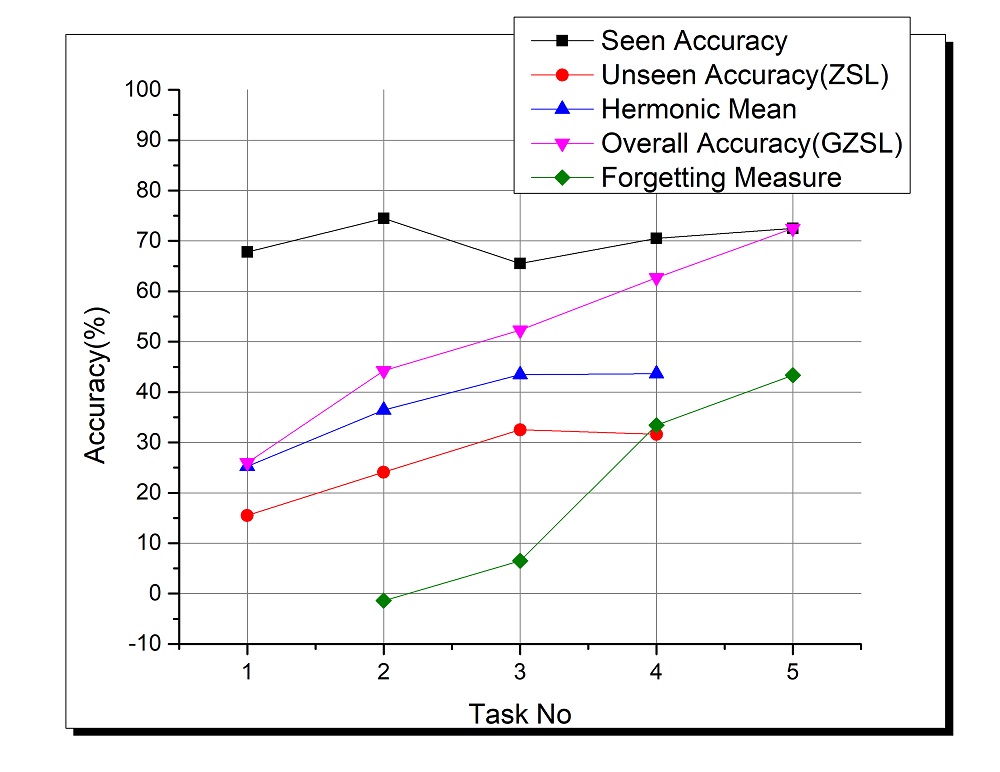}
    \caption{Results with adversarial training for the AWA2 dataset.}
    \label{fig:b10}
\end{figure}

We divide the aPY dataset into 5 tasks, where each consists of ten classes. We compare our results with several methods in Table \textcolor{blue}{[\ref{table:5}]}. Generalized CZSL\textcolor{blue}{\cite{a6}} used memory to store real data as the replay, and it achieved mH = 38.34. In contrast, our model achieves mH = 37.19 when each task gets trained for 100 epochs. Similarly, the classifier gets trained for 30 epochs for all tasks. The performance of our model per task is given in Figures \textcolor{blue}{[\ref{fig:b9}, \ref{fig:b10}]}.

\section{Conclusion}
In this work, we proposed a novel hybrid algorithm. The novelty of our work is that we use adversarial learning. Here the model needs generative replay and it grows for task incremental learning. The private module barely shares knowledge from seen to unseen classes that can be future work to optimize the private module for ZSL. Another future work might be to develop task-free continual zero-shot learning. How can we build a continual zero-shot learning model for object detection? What should be the optimum latent dimension, hidden-layer size?
\section*{Acknowledgement}
    We thank prof Suresh Sundaram and Dr. Chandan Gautam from the Artificial Intelligence(AI) lab in the Aerospace Engineering Department, Indian Institute of Science(IISc), Bangalore, for the valuable discussion in the initial phase of this work. 
\newpage

\bibliographystyle{plain}
\bibliography{egbib}

\end{document}